%% file: main.tex
\documentclass{article}

\usepackage[nonatbib,preprint]{neurips_2023_modified}
\usepackage[numbers,sort&compress]{natbib}

\input{macros}

\title{Large Language Models for In-Context Student Modeling: Synthesizing Student's Behavior in Visual Programming}

\author{%
  Manh Hung Nguyen\\
  MPI-SWS, Germany\\
  \text{manguyen@mpi-sws.org} \\
  \And
  Sebastian Tschiatschek \\
  University of Vienna, Austria \\
  \text{sebastian.tschiatschek@univie.ac.at} \\
  \And
  Adish Singla \\
  MPI-SWS, Germany \\
  \text{adishs@mpi-sws.org} \\
}

\begin{document}

\maketitle

\input{0_abstract}
\input{1_introduction}
\input{2_relatedwork}
\input{3_problemsetup}
\input{4_methodology}
\input{5_experiments}

\input{6_conclusions}
\input{7_acknowledgements}
\bibliographystyle{unsrt}
\bibliography{main}

\end{document}

%% file: macros.tex
\usepackage[utf8]{inputenc} 
\usepackage[T1]{fontenc}    
\usepackage{url}            
\usepackage{booktabs}       
\usepackage{amsfonts}       
\usepackage{nicefrac}       
\usepackage{microtype}      

\usepackage[usenames,dvipsnames]{xcolor}
\usepackage{hyperref}
\hypersetup{ %
	pdfauthor={},
	pdftitle={},
	pdfsubject={},
        pdflang={en},
	bookmarks,
	bookmarksnumbered,
	backref=page,
	pageanchor=true,
	plainpages=false,
	colorlinks=true,%
	linktocpage=true,
	citecolor=MidnightBlue,%
	filecolor=BrickRed,%
	linkcolor=NavyBlue,%
	urlcolor=NavyBlue 
}

\usepackage{framed}
  {\endMakeFramed}
\usepackage{svg}
\usepackage{subcaption}
\usepackage[most]{tcolorbox}
\usepackage{multirow}
\usepackage{enumerate}   
\usepackage{boxedminipage}
\usepackage{listings}
\usepackage{graphicx}
\usepackage{colortbl}
\usepackage{xspace}
\usepackage[inline]{enumitem}
\usepackage{adjustbox}
\usepackage{tabularx} 
\usepackage{fancybox} 

\lstdefinestyle{mystyle}{
    commentstyle=\color{gray},
    keywordstyle=\color{blue},
    numberstyle=\tiny\color{gray},
    stringstyle=\color{purple},
    basicstyle=\ttfamily\fontsize{9}{9}\linespread{0.8}\selectfont,
    breakatwhitespace=false,         
    breaklines=true,                 
    captionpos=b,                    
    keepspaces=true,                 
    numbers=left,                    
    numbersep=3pt,                  
    showspaces=false,                
    showstringspaces=false,
    showtabs=false,                  
    tabsize=2,
    belowskip=0pt,
    escapechar=\&
}
\usepackage{ifthen}
\newboolean{include-notes}
\setboolean{include-notes}{true}
\newcommand{\nb}[3]{\ifthenelse{\boolean{include-notes}}{{\colorbox{#2}{\bfseries\sffamily\scriptsize\textcolor{white}{#1}}}{\ \textcolor{#2}{\sf\small\textit{#3}}}}{}}

\definecolor{ExpHighlight}{rgb}{0.8, 0.1, 0.1}
\definecolor{mygreen}{rgb}{0,0.6,0}
\definecolor{mygray}{rgb}{0.5,0.5,0.5}
\definecolor{mymauve}{rgb}{0.58,0,0.82}
\definecolor{CodeHighlight}{rgb}{1,1,0.6}
\definecolor{CodeGray}{rgb}{0.8,0.8,0.8}
\definecolor{CodeDarkGray}{rgb}{0.6,0.6,0.6}
\definecolor{InputOutput}{rgb}{0.764,0.345,0.009}
\definecolor{promptinputcolor}{rgb}{0.16, 0.32, 0.75}
\definecolor{promptheadercolor}{rgb}{0.764,0.345,0.009}
\definecolor{highlightrow}{rgb}{0.886,1,0.772} 

\newcommand{\oursrowcolor}{green!15}

\newcommand{\reference}{\ensuremath{\textnormal{ref}}}

\newcommand{\student}{\ensuremath{\textsc{stu}}}
\newcommand{\target}{\ensuremath{\textnormal{tar}}}
\newcommand{\tar}{\target}

\newcommand{\ft}{\ensuremath{\textnormal{ft}}}
\newcommand{\textcode}[1]{{\fontfamily{cmtt}\selectfont #1}\xspace}
\newcommand{\studentsyn}{\textbf{\resizebox{!}{0.95em}{\MakeUppercase{S}}\resizebox{!}{0.7em}{TUDENT}\resizebox{!}{0.95em}{\MakeUppercase{S}}\resizebox{!}{0.7em}{YN}}}
\newcommand{\neurss}{$\!$\textbf{\resizebox{!}{0.75em}{\MakeUppercase{N}}\resizebox{!}{0.58em}{EUR}\resizebox{!}{0.75em}{\MakeUppercase{SS}}}$\!$}

%% file: 0_abstract.tex
\begin{abstract}
Student modeling is central to many educational technologies as it enables predicting future learning outcomes and designing targeted instructional strategies. However, open-ended learning domains pose challenges for accurately modeling students due to the diverse behaviors and a large space of possible misconceptions. To approach these challenges, we explore the application of large language models (LLMs) for in-context student modeling in open-ended learning domains. More concretely, given a particular student's attempt on a reference task as observation, the objective is to synthesize the student's attempt on a target task. We introduce a novel framework, \emph{LLM for Student Synthesis} (\textsc{LLM-SS}), that leverages an LLM for synthesizing a student's behavior. Our framework can be combined with different LLMs; moreover, we fine-tune LLMs to boost their student modeling capabilities. We instantiate several methods based on \textsc{LLM-SS} framework and evaluate them using an existing benchmark, \textsc{StudentSyn}, for student attempt synthesis in a visual programming domain. Experimental results show that our methods perform significantly better than the baseline method \textsc{NeurSS} provided in the \textsc{StudentSyn} benchmark. Furthermore, our method using a fine-tuned version of the GPT-3.5 model is significantly better than using the base GPT-3.5 model and gets close to human tutors' performance.
\end{abstract}

%% file: 1_introduction.tex
\section{Introduction}
\textit{Student modeling} refers to the process of representing the current state of a learner's knowledge, skills, preferences, and learning needs \cite{vanlehn2013student}. This is pivotal in developing educational systems as it allows for the personalization of learning experiences \cite{Chrysafiadi2015SM}, catering specifically to each student's unique abilities and growth areas, and targeted instructional strategies that can significantly enhance the learning process \cite{DBLP:conf/edm/RaffertyJG16}. By understanding student behavior, tutoring systems and educators can identify patterns and trends \cite{DBLP:conf/sigcse/EmersonSRWMBL20, DBLP:conf/lak/0004SWMPP21}, thereby predicting future learning outcomes \cite{DBLP:conf/edm/WangSLP17} and providing timely support. Moreover, it allows them to detect if and when a student is losing interest or facing challenges \cite{DBLP:conf/edm/CockMGK21}, enabling them to intervene effectively~\cite{DBLP:conf/aied/GhoshTDS22}. In particular, student modeling is key in \textit{open-ended learning domains} where creativity and exploratory behaviors are encouraged \cite{Hannafin1994ET,DBLP:journals/aiedu/KaserS20}

\input{figs/problemsetup/fig_problemsetup}

In open-ended learning domains such as programming, students can take different learning paths and complete a task with different strategies \cite{Hannafin1994ET}. This results in diverse behaviors and presents significant challenges to modeling a particular student's behavior \cite{DBLP:journals/aiedu/KaserS20}. In recent years, some efforts in student modeling for open-ended learning domains have been made, such as representing knowledge and forecasting future performance using deep learning \cite{DBLP:conf/edm/WangSLP17}, investigating students' problem-solving approaches using Natural Language Processing \cite{DBLP:conf/lak/KongHBO23}, early prediction of conceptual understanding \cite{DBLP:conf/edm/CockMGK21}, clustering-based methods for misconception discovery \cite{DBLP:conf/sigcse/EmersonSRWMBL20}, students' attempts synthesis in block-based visual programming \cite{DBLP:journals/corr/abs-2205-01265}, and predicting students' post-test performance and interest using multimodal predictive student modeling \cite{DBLP:conf/lak/EmersonMRAL23}. Existing works on student modeling in open-ended learning domains often require a large behavioral dataset from students or use a complex pipeline, and sometimes, a combination of both \cite{DBLP:conf/edm/WangSLP17, DBLP:journals/aiedu/FratamicoCKR17, DBLP:journals/corr/abs-2205-01265, DBLP:conf/lak/EmersonMRAL23}. In this paper, we seek to leverage recent advances in generative AI and large language models (LLMs) for student modeling in open-ended learning domains and address the above-mentioned shortcomings.

In particular, LLMs have demonstrated advanced capabilities for \textit{in-context learning} in which a model learns to solve a downstream application scenario when prompted with appropriate contextual information~\cite{DBLP:conf/nips/BrownMRSKDNSSAA20,DBLP:journals/corr/abs-2303-12712}. Notably, they have been used to simulate humans for replicating human subject studies \cite{pmlr-v202-aher23a} and to simulate students for training teaching assistants \cite{DBLP:conf/lats/MarkelOLP23}. In this work, we investigate the potential of leveraging such capabilities of LLMs for \textit{in-context student modeling} in open-ended learning environments.
In our setup, an LLM observes a student's attempt on a reference task as the student's behavioral context, and the objective is to synthesize the student's attempt on a target task, reflecting the student's problem-solving style and misconceptions observed. In essence, we seek to address the following research question: \textit{Given a specific student's behavioral context, are LLMs capable of effectively modeling the student and subsequently synthesizing the student's attempt on a target task?}

To this end, we introduce a novel framework, \emph{LLM for Student Synthesis} (\textsc{LLM-SS}), that leverages LLMs for modeling and synthesizing a student's behavior. The design of our framework is inspired by \emph{Perturbation Student Model} \cite{Kass89Student}, based on the idea that a student's knowledge can be modeled as perturbations to expert knowledge. Our framework operationalizes this idea by providing a student's behavioral context in the prompt and improving the expert knowledge of a base LLM via fine-tuning. In summary, our main contributions are:
\begin{enumerate}[label=\Roman*.,leftmargin=20pt,parsep=1.5pt]

    \item We formalize the problem of using an LLM's in-context learning capabilities for student modeling and behavior synthesis in open-ended learning domains. 
    \item We propose a novel framework \textsc{LLM-SS} for synthesizing student's behavior. Our framework can be combined with different LLMs; moreover, we fine-tune LLMs to boost their student modeling capabilities.
    \item We evaluate several methods instantiated from our framework on an existing benchmark, \textsc{StudentSyn},  for student attempt synthesis in a visual programming domain. Our results highlight that our methods perform significantly better than baselines without requiring complex pipelines or extensive datasets.
    \item We publicly release the implementation of \textsc{LLM-SS} to facilitate future research.\footnote{Github: \url{https://github.com/machine-teaching-group/edm2024-llm-student-modeling}}
\end{enumerate}

%% file: figs/problemsetup/fig_problemsetup.tex
\begin{figure*}
    \centering
    \includegraphics[width=1\textwidth]{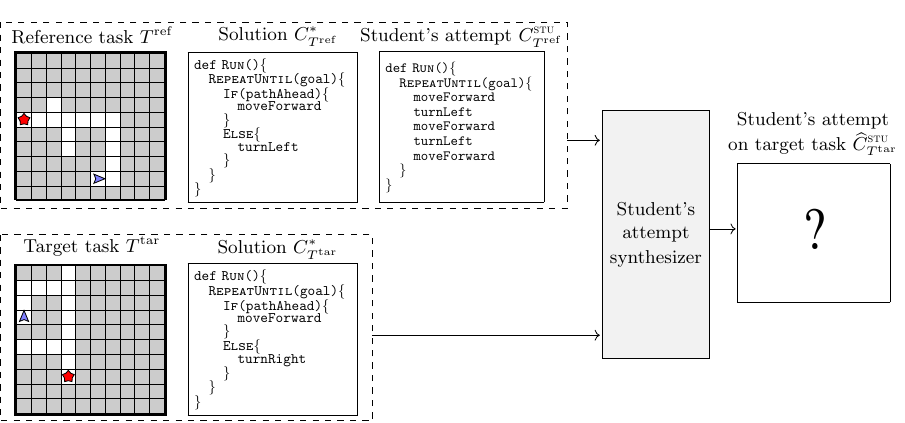}
    \vspace{0.1cm}
    \caption{Illustration of our problem setup in a visual programming environment. The scenario is taken from the \textsc{StudentSyn} benchmark \cite{DBLP:journals/corr/abs-2205-01265}.
    A synthesizer observes a tuple of ($T^{\reference}$, $C^*_{T^{\reference}}$, $C^{\student}_{T^{\reference}}$) indicating a student \student's behavior. Then, given a target task $T^{\tar}$ along with a solution $C^*_{T^{\tar}}$, the synthesizer generates a student's attempt $\widehat{C}^{\student}_{T^{\tar}}$ that imitates the student's behavior.}
    \label{fig:problem-setup}
\end{figure*}

%% file: 2_relatedwork.tex

\section{Related Work}
\textbf{Student modeling and synthesis in open-ended domains.} As discussed in the previous section, there have been recent developments on student modeling for open-ended learning domains, with techniques ranging from misconception discovery to identification of struggling students and investigating problem-solving strategies~\cite{DBLP:conf/sigcse/EmersonSRWMBL20,DBLP:conf/lak/0004SWMPP21,DBLP:conf/edm/CockMGK21,DBLP:conf/edm/WangSLP17, DBLP:conf/lak/EmersonMRAL23,DBLP:journals/corr/abs-2205-01265,DBLP:conf/lak/KongHBO23}. Among these recent works, our work is closer to that of \cite{DBLP:journals/corr/abs-2205-01265} as we are addressing the problem of synthesizing a student's behavior by focusing on misconceptions in observed attempts. In fact, our evaluation is based on the \textsc{StudentSyn} benchmark from \cite{DBLP:journals/corr/abs-2205-01265} that considers the problem of synthesizing a student's attempt in visual programming domains. As part of this benchmark, \cite{DBLP:journals/corr/abs-2205-01265} proposed an automated method, \textsc{NeurSS}, that requires extensive pre-training on expert data and continual training on real-world data from similar students. Our framework aims to avoid this complex training pipeline by leveraging the in-context learning capabilities of LLMs. Our work is also similar in spirit to contemporary works that use LLMs for simulating students to teach learners in conversational tutoring systems~\cite{neurips2023gaied_38_schmucker} or train human tutors \cite{DBLP:conf/lats/MarkelOLP23}.

\textbf{LLMs in programming education.} Generative AI and LLMs hold great promise in enhancing the field of education through a complementary relationship between human teachers and generative models~\cite{DBLP:journals/corr/abs-2402-01580,DBLP:journals/eait/JeonL23a}. Some of the earlier works applying LLMs in educational settings focused on computing and programming education domains and studied a various of scenarios, including generating high-precision feedback \cite{DBLP:journals/corr/abs-2302-04662,GPT4Hints-GPT3.5Val}, generating programming exercises \cite{DBLP:conf/icer/SarsaDH022}, repairing bugs in programming assignments~\cite{DBLP:journals/corr/abs-2209-14876}, task synthesis for visual programming \cite{neurtasksyn,DBLP:conf/icer/Singla22}, and benchmarking LLMs capabilities with that of human tutors~\cite{DBLP:conf/icer/Singla22,DBLP:conf/icer/PhungPCGKMSS22}. Our work differs from these works, given our focus on leveraging LLMs for modeling a student and synthesizing students' attempts.

%% file: 3_problemsetup.tex

\section{Problem Setup}
\label{sec:problem-setup}

In this section, we formalize the problem of leveraging LLMs for in-context student modeling in open-ended domains. While we focus on LLM-based methods, we provide a generic setup that encapsulates various baseline methods that do not use LLMs (e.g., baseline \textsc{NeurSS} used for comparison in Section~\ref{sec:experiment}). In particular, our problem setup is inspired by the work of \cite{DBLP:journals/corr/abs-2205-01265} that will also be used later as a benchmark in our experiments.

\textbf{Preliminaries and synthesis objective.} Given an open-ended learning domain, there is a student, henceforth referred to as \student, aiming to solve some tasks in the domain. We denote the space of all possible tasks by $\mathbb{T}$, and the space of all possible solutions and attempts by $\mathbb{C}$. In particular, we are given a reference task $T^{\reference} \in \mathbb{T}$ of interest along with a solution $C_{T^{\reference}}^* \in \mathbb{C}$.\footnote{A task $T$ can have multiple solutions, and $C^*_{T}$ refers to any solution codes written by experts being provided as input.} 
Our main goal is to develop a synthesizer that can model the student $\,\student\,$ by observing how  $\,\student\,$  solves $T^{\reference}$, and subsequently synthesize an attempt on any similar target task $T^{\target}$, imitating \student's behavior. More concretely, we consider the following two-step process:
\begin{enumerate}[label=(\arabic*),leftmargin=*,parsep=1.5pt]
  \item First, the synthesizer observes a student's \textit{context} tuple ($T^{\reference}$, $C^*_{T^{\reference}}$, $C^{\student}_{T^{\reference}}$), where $C^{\student}_{T^{\reference}} \in \mathbb{C}$ is the student \student's attempt on solving the reference task.  
  \item Next, given a target task $T^{\tar} \in \mathbb{T}$ conceptually similar to $T^{\reference}$, along with a solution $C_{T^{\tar}}^* \in \mathbb{C}$, the synthesizer synthesizes a student's attempt $\widehat{C}$$^{\student}_{T^{\tar}}$, which should be close to how the student $\student$ would attempt $T^{\tar}$.\footnote{There are different granularity levels at which we can synthesize the student $\,\student\,$'s behavior, including: (a) a coarse-level binary prediction of success/failure, (b) a medium-level prediction w.r.t.\ predefined misconceptions; (c) a fine-level synthesis of student's attempt. Here, we focus on this fine-level objective of synthesizing a student's attempt.}
\end{enumerate}

\textbf{Quality rubric for evaluation.} We evaluate the performance of a synthesizer based on the quality of their synthesized student's attempt $\widehat{C}$$^{\student}_{T^{\tar}}$. Based on existing literature~\cite{DBLP:journals/corr/abs-2205-01265,DBLP:conf/icer/PhungPCGKMSS22}, we quantitatively measure the generative quality using expert-based assessments w.r.t.\ the following quality rubric:
\begin{itemize}[leftmargin=*,parsep=1.5pt]
    \item \textsc{Q-\student}. This attribute measures whether the synthesized attempt $\widehat{C}^{\student}_{T^{\tar}}$ captures the student \student's behavior (e.g., problem-solving strategy and underlying misconceptions).
    \item \textsc{Q-task}. This attribute measures whether the synthesized attempt $\widehat{C}^{\student}_{T^{\tar}}$ captures the characteristics of $T^{\tar}$ (e.g., partially reflecting its solution $C^*_{T^{\tar}}$).
    \item \textsc{Q-overall}. This attribute measures whether the synthesized attempt $\widehat{C}^{\student}_{T^{\tar}}$ successfully captures both the student's behavior and the target task's characteristics at the same time. We will set \textsc{Q-overall} $=$ \textsc{Q-\student} $\times$ \textsc{Q-task}.
\end{itemize}

\textbf{Illustrative example for visual programming domain.} In our experimental evaluation (Section~\ref{sec:experiment}), we will consider an existing benchmark, \textsc{StudentSyn}~\cite{DBLP:journals/corr/abs-2205-01265}, for student attempt synthesis in a visual programming domain of \textit{Hour of Code: Maze Challenge by Code.org} (HoCMaze)~\cite{hourofcode_maze}. As an illustrative example, Figure~\ref{fig:problem-setup} shows a concrete scenario for our problem setup.

%% file: 4_methodology.tex
\section{Our LLM-SS Framework}
\label{sec:llm-ss}
In this section, we propose a novel framework, namely \textsc{LLM-SS}, for in-context student modeling and synthesizing students' attempts. It is inspired by the \emph{Perturbation Student Model} as discussed below (Section \ref{sec:perturbation-student-model}). Afterward, we delve into two components of LLM-SS: providing student's context (Section \ref{sec:main-prompt}) and providing domain expertise (Section \ref{sec:fine-tune-llm}).    

\subsection{Perturbation Student Model}
\label{sec:perturbation-student-model}

\emph{Perturbation Student Model} is based on the idea that a student's knowledge can be modeled as perturbations to expert knowledge \cite{Kass89Student}. This model was introduced as an extension of the \emph{Overlay Model}~\cite{carr:hal-00702959,Greer94Student} -- it allows modeling a student's misconceptions and ``buggy'' knowledge that deviates from expert knowledge. It assumes that incorrect behaviors of a student can be caused by systematically applying a set of perturbations to domain expertise. 

In our LLM-SS framework, we use an LLM to model a student in an open-ended learning domain following the same idea of \emph{Perturbation Student Model}. More concretely, we provide a student's knowledge by a behavioral context in a prompt to LLM (Section \ref{sec:main-prompt}), and provide domain-specific expertise through fine-tuning the LLM on expert data (Section \ref{sec:fine-tune-llm}).  

\subsection{Providing Student's Context}
\label{sec:main-prompt}
Next, we discuss how to provide a student's context to an LLM and leverage the LLM's capabilities of in-context learning. Again, the goal of a student's context is to give an LLM information about the student, which may include the student's background, preferences, learning history, and problem-solving trajectories on multiple tasks. This information can be provided to a given LLM as a context in a prompt -- existing works have shown that LLMs can effectively learn from such contextual information without explicit training or further parameter updates \cite{DBLP:conf/nips/BrownMRSKDNSSAA20,DBLP:journals/corr/abs-2303-12712}. 

In our framework, the prompt includes a student's context in the form of a problem-solving attempt on a reference task, which is represented by an information tuple ($T^{\reference}$, $C^*_{T^{\reference}}$, $C^{\student}_{T^{\reference}}$); see Section~\ref{sec:problem-setup}. We expect the LLM to infer the student's misconceptions from the observed attempt along with the necessary perturbations to obtain $C^{\student}_{T^{\reference}}$ from $C^*_{T^{\reference}}$. Subsequently, the LLM is asked to play the role of this student and synthesize an attempt for a target task $T^{\target}$, which should reflect the student's behavior. This is when the LLM should apply the same perturbations to obtain $\widehat{C}^{\student}_{T^{\target}}$ from $C^*_{T^{\target}}$. 

Figure~\ref{fig:main-prompt} shows an example of our main prompt template for providing the student \student's context and synthesizing the student's attempts. We note that our \textsc{LLM-SS} framework can accommodate multiple solutions for a task and richer representations of the student's context as input by appropriately adapting the prompt. In this template example, we have shown a single solution for a task and one student's attempt, as considered in our experimental evaluation; see Section~\ref{sec:experiment}. 

\input{figs/methodology/fig_prompt}
\input{figs/methodology/fig_fine-tuning}

\subsection{Providing Domain Expertise}
\label{sec:fine-tune-llm}

Next, we discuss how to provide domain-specific expertise to an LLM for student modeling. In general, datasets used for pre-training LLMs may not contain data coming from specialized open-ended learning domains such as interactive educational games \cite{DBLP:journals/aiedu/KaserS20},  physics simulations \cite{doi:10.1126/science.1161948}, or visual programming \cite{hourofcode_maze}. Consequently, LLMs could be far from experts in these domains; for instance, even state-of-the-art models like GPT-4 perform poorly in synthesizing solutions for visual programming tasks~\cite{DBLP:conf/icer/Singla22}. In such settings, we need to enhance an LLM domain-specific knowledge to effectively model a student as per the \emph{Perturbation Student Model}. In particular, we will enhance an LLM's domain expertise via fine-tuning -- existing works have shown that pre-trained LLMs can be tailored to specific domains via fine-tuning~\cite{DBLP:conf/nips/Ouyang0JAWMZASR22, DBLP:journals/corr/abs-2307-09288}

In our framework, we aim to improve a given LLM's capability of generating solutions $C^*_T$ for any task $T$ similar to the reference task $T^{\reference}$. Once the LLM acquires a better understanding of how to solve tasks in the domain, it is expected to better infer the student's behavior from the context provided in Section~\ref{sec:main-prompt}. More concretely, we use pairs of (task $T$, solution $C^*_T$) in the domain to create a fine-tuning dataset $\mathbb{D}_\ft = \{\mathbf{x}^{(k)},\mathbf{y}^{(k)}\}$, where $\mathbf{x}^{(k)}$ is an input prompt containing a task to be solved and $\mathbf{y}^{(k)}$ is the desired solution generated by the LLM. We consider an LLM parameterized by $\theta$, with $p_\theta$ denoting conditional probability distribution of sampling responses. We perform supervised fine-tuning to adjust $\theta$ through gradient descent, with the objective of minimizing the negative log-likelihood loss given by $L_{\ft}(\theta) := - \mathbb{E}_{(\mathbf{x}^{(k)},\mathbf{y}^{(k)}) \sim \mathbb{D}_{\ft}}\bigl[\textnormal{log}~ p_\theta(\mathbf{y}^{(k)}|\mathbf{x}^{(k)})\bigl]$~\cite{DBLP:journals/corr/abs-2307-09288}.

\looseness-1Figure~\ref{fig:fine-tuning} shows the pipeline overview of fine-tuning an LLM in our framework along with an example of fine-tuning prompt template. In each prompt $\mathbf{x}^{(k)}$, we first start by describing the domain background (same as in Figure~\ref{fig:main-prompt}). Then, we use an instruction to steer the LLM's behavior to act as a domain expert and solve a task. The last part of the prompt is a representation of the task to be solved. 

%% file: figs/methodology/fig_prompt.tex

\begin{figure}[t!]
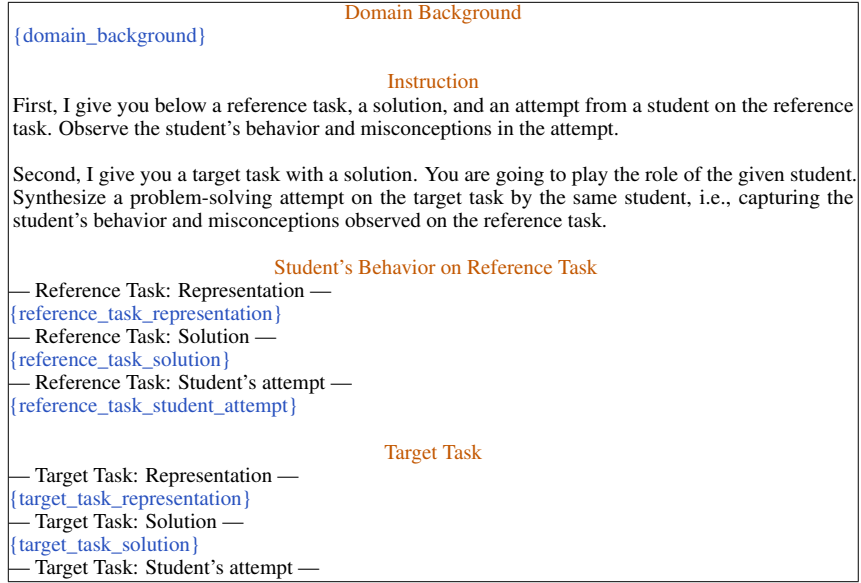
    
    \centering
    \scalebox{0.8}{
    \setlength\tabcolsep{2pt}
    \renewcommand{\arraystretch}{1.0}
        \begin{tabular}{|p{1.00\linewidth}|}
        \hline
        \centerline{\color{promptheadercolor}\textnormal{Domain Background}}
            {\color{promptinputcolor}\{domain\_background\}}\newline 
            
            \centerline{\color{promptheadercolor}\textnormal{Instruction}}
            
            First, I give you below a reference task, a solution, and an attempt from a student on the reference task. Observe the student's behavior and misconceptions in the attempt.\newline
        
            Second, I give you a target task with a solution. You are going to play the role of the given student. Synthesize a problem-solving attempt on the target task by the same student, i.e., capturing the student's behavior and misconceptions observed on the reference task.\newline
            
            \centerline{\color{promptheadercolor}\textnormal{ Student's Behavior on Reference Task}}
    
            --- Reference Task: Representation ---

            {\color{promptinputcolor}\{reference\_task\_representation\}}
            
            --- Reference Task: Solution ---
            
            {\color{promptinputcolor}\{reference\_task\_solution\}}
            
            --- Reference Task: Student's attempt ---
            
            {\color{promptinputcolor}\{reference\_task\_student\_attempt\}}\newline

            \centerline{\color{promptheadercolor}\textnormal{Target Task}}
            
            --- Target Task: Representation ---
            
            {\color{promptinputcolor}\{target\_task\_representation\}} 
            
            --- Target Task: Solution ---
            
            {\color{promptinputcolor}\{target\_task\_solution\}} 
            
            --- Target Task: Student's attempt ---
        \\
        \hline
        \end{tabular}
    }
    \caption{Prompt template used in LLM-SS framework. {\color{promptinputcolor}\{placeholders\}} are used to include details for each scenario.}
    \vspace{-8mm}
    \label{fig:main-prompt}
\end{figure}

%% file: figs/methodology/fig_fine-tuning.tex

\begin{figure}[t!]
   \begin{subfigure}[b]{0.45\textwidth}
        \centering
        \scalebox{0.9}{
        \setlength\tabcolsep{2pt}
        \renewcommand{\arraystretch}{1.0}
            \begin{tabular}{|p{1.0\linewidth}|}
                \hline
                \centerline{\color{promptheadercolor}\textnormal{Domain Background}}
                {\color{promptinputcolor}\{domain\_background\}}\newline 
                
                \centerline{\color{promptheadercolor}\textnormal{Instruction}}
                You should act as an expert in this domain and synthesize a solution for the following task below.\newline

                \centerline{\color{promptheadercolor}\textnormal{Task}}
                
                --- Task: Representation ---
                
                {\color{promptinputcolor}\{task\_representation\}} 
                
                --- Task: Solution ---
                \\
                \hline
            \end{tabular}
        }
        \subcaption{\centering Prompt for fine-tuning.}
        \label{fig:fine-tuning-prompt}
   \end{subfigure} 
   \hfill
    \begin{subfigure}[b]{0.55\textwidth}
        \centering
        \includegraphics[trim=1em 0em 0em 0em, clip, width=1\textwidth]{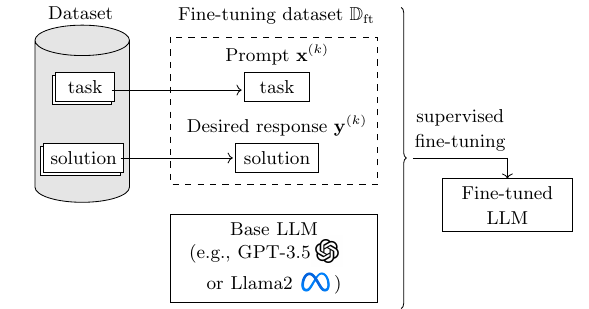}
        \subcaption{\centering{Fine-tuning an LLM using synthetic data.}}
        \label{fig:fine-tuning-pipeline}
   \end{subfigure}
    \caption{Fine-tuning an LLM using expert knowledge in \textsc{LLM-SS} framework.}
    \label{fig:fine-tuning}
\end{figure}

%% file: 5_experiments.tex

\section{Experimental Evaluation}
\label{sec:experiment}
\vspace{-2mm}
This section presents our experimental evaluation, including description of \textsc{StudentSyn} benchmark with baseline methods from \cite{DBLP:journals/corr/abs-2205-01265} (Section \ref{sec:StudentSyn-benchmark}), evaluated methods (Section \ref{sec:our-models}), evaluation procedure (Section \ref{sec:eval-procedure}), and results (Section \ref{sec:results}).
\vspace{-3mm}
\subsection{\studentsyn ~Benchmark and Baselines}
\label{sec:StudentSyn-benchmark}

We use the \textsc{StudentSyn} benchmark from \cite{DBLP:journals/corr/abs-2205-01265}, designed to evaluate student's attempt synthesis methods in the visual block-based programming domain of \textit{Hour of Code: Maze Challenge by Code.org} (HoCMaze) \cite{hourofcode_maze}. This programming domain has been popularly used in several existing works \cite{DBLP:conf/lats/PiechSHG15,DBLP:conf/edm/EfremovGS20,DBLP:conf/nips/AhmedCEFGRS20,DBLP:journals/corr/abs-2205-01265}. Figure \ref{fig:problem-setup} shows an example of task $T^{\reference}$ along with a solution $C^*_{T^{\reference}}$ -- a task in HoCMaze is specified by a visual grid containing an avatar (blue arrow), a goal (red star), and some walls (gray cells); a solution code brings the avatar to the goal's location while avoiding hitting the walls. \textsc{StudentSyn} is a challenging benchmark for our problem setup, as evidenced by the huge performance gap between human tutors and automated methods proposed in  \cite{DBLP:journals/corr/abs-2205-01265}.

\textbf{Benchmark scenarios.} This benchmark comprises two reference tasks $T^{\reference}$, namely HoCMaze-4 and HoCMaze-18~\cite{hourofcode_maze}, and three target tasks $T^{\target}$ associated with each reference task. In our illustration of problem setup in Figure \ref{fig:problem-setup}, we use HoCMaze-18 as $T^{\reference}$. The benchmark considers six types of misconceptions, such as confusion between left/right directions when turning, writing repetitive turn commands, and ignoring the If-Else/While structure. The benchmark provides a set of scenarios comprising a student $\student\!$ with a specific misconception, one coding attempt $C^{\student}_{T^{\reference}}$ on each reference task, and the \student's attempt on each target task $C^{\student}_{T^{\target}}$ serving as a ground-truth. In total, we evaluate on these $36$ scenarios ( $2 \ T^{\reference} \times 3 \ T^{\target} \times 6 \ \student$). 

\looseness-1\textbf{Dataset for fine-tuning.} Along with benchmark scenarios, \cite{DBLP:journals/corr/abs-2205-01265} also provides a synthetic dataset consisting of (task, solution) pairs, where tasks are similar to either HoCMaze-4 or HoCMaze-18. Here, task similarity is measured conceptually by edit distance in solution codes. This synthetic dataset was created and used for pre-training models introduced in \cite{DBLP:journals/corr/abs-2205-01265}. In our framework, we will use it to fine-tune a base LLM to boost its domain expertise. In total, there are $10,000$ training tasks and $500$ validation tasks for HoCMaze-4, and $40,000$ training tasks and $500$ validation tasks for HoCMaze-18.

\textbf{Baseline methods.} We compare our framework with baseline method \textsc{NeurSS} \cite{DBLP:journals/corr/abs-2205-01265}, an LSTM-based neural network pre-trained on expert knowledge and continually trained on real students' attempts. We also compare our framework with human tutors in the visual programming domain, referred to as \textsc{TutorSS} in \cite{DBLP:journals/corr/abs-2205-01265}. Here, \textsc{TutorSS} can be considered an oracle that provides performance upper bounds. We re-use the students' attempts synthesized by \textsc{NeurSS} and \textsc{TutorSS} from \cite{DBLP:journals/corr/abs-2205-01265}, and re-assess them w.r.t. our rubric in Section \ref{sec:eval-procedure}.

\vspace{-2mm}
\subsection{Methods Based on LLM-SS Framework}
\label{sec:our-models}

\textbf{Methods using a base LLM without fine-tuning.} Based on our \textsc{LLM-SS} framework, we develop the following concrete methods using base models without fine-tuning step described in Section \ref{sec:fine-tune-llm}: GPT-3.5-SS using GPT-3.5 \cite{GPT3.5}, GPT-4-SS using GPT-4 \cite{DBLP:journals/corr/abs-2303-08774}, Llama2-7B-SS using Llama2-7B-Chat \cite{DBLP:journals/corr/abs-2307-09288}, and Llama2-70B-SS using Llama2-70B-Chat \cite{DBLP:journals/corr/abs-2307-09288}.

\textbf{Methods using a fine-tuned LLM.} We further develop the following three concrete methods by fine-tuning three models: GPT-3.5ft-SS using fine-tuned GPT-3.5 \cite{GPT3.5}, Llama2-7Bft-SS using fine-tuned Llama2-7B-Chat \cite{DBLP:journals/corr/abs-2307-09288}, and Llama2-70Bft-SS using fine-tuned Llama2-70B-Chat \cite{DBLP:journals/corr/abs-2307-09288}. We did not fine-tune the GPT-4 model as APIs are not publicly available. Details of our fine-tuning procedure are explained in Section \ref{sec:fine-tune-llm}.\footnote{For fine-tuning Llama2-70B models, we used a cluster of $2\times36$ cores, $2.40$ GHz Intel Xeon Platinum Processor $8360$Y, and $8\times$Nvidia A$100$ $80$GB, with parallelization under a $64$-bit Debian. We fine-tuned a model for each reference task separately, and one run on a reference task took up to $35$ hours. For GPT-3.5, we fine-tuned the GPT-3.5-turbo-0613 model for each reference task separately, and one run on a reference task took up to $7$ hours. We paid about $1000$\$ in total for using fine-tuning APIs provided by OpenAI.}

\subsection{Evaluation Procedure}
\label{sec:eval-procedure}

For each scenario from the \textsc{StudentSyn} benchmark (see Section \ref{sec:StudentSyn-benchmark}), we create a prompt following the template in Figure~\ref{fig:main-prompt} to use it as input to an LLM. We use the domain background representation for HoCMaze based on prompts in recent works~\cite{neurtasksyn}. Subsequently, all scenarios together with student attempts synthesized by the LLM are presented to two independent experts for assessment -- these two experts have extensive knowledge in computer science and visual programming domains, and follow the evaluation rubric from Section~\ref{sec:problem-setup}. The annotation process is done in a blind condition, in which experts do not know from which method a coding attempt is synthesized. In total, there are about $500$ codes to be annotated by each expert corresponding to different scenarios and methods.

\input{figs/experiments/fig_results}

These two experts annotated the synthesized codes by using binary values $\{0, 1\}$ for annotation, i.e., each quality attribute could take a value of $0$ (bad) or $1$ (good). Concretely, $\textsc{Q-stu}=1$ means that $\widehat{C}^{\student}_{T^{\tar}}$ captures the student \student's behavior in terms of the problem-solving strategy and underlying misconceptions, and otherwise $\textsc{Q-stu}=0$; similarly, $\textsc{Q-task}=1$ means that $\widehat{C}^{\student}_{T^{\tar}}$ captures the characteristics of target task $T^\target$, and otherwise $\textsc{Q-task}=0$. \textsc{Q-overall}, defined as \textsc{Q-\student} $\times$ \textsc{Q-task}, takes values of $\{0, 1\}$. We validate the expert annotations w.r.t. \textsc{Q-overall} using Cohen's kappa inter-agreement reliability~\cite{Cohen1960ACO}, obtaining a value of $0.71$, indicating \emph{substantial agreement} between two experts.

Nevertheless, further investigation into the annotations revealed that the majority of disagreements between two experts were borderline cases where the quality attribute value was unclear. This motivated us to refine the scale of assessment where \textsc{Q-\student} and \textsc{Q-task} would take values of $\{0, 0.5, 1\}$, with $0.5$ now indicating partially capturing the student's behavior or the target task's characteristics. We note that \textsc{Q-overall},  defined as \textsc{Q-\student} $\times$ \textsc{Q-task}, now takes values of $\{0, 0.25, 0.5, 1\}$. With this refined scale, one expert did the entire annotations again and the final results reported in Section~\ref{sec:results} are based on these new annotations. We report averaged results in the range $[0.0, 1.0]$ by aggregating across all scenarios for a given reference task.

\subsection{Results}
\label{sec:results}

\input{figs/experiments/fig_loss}
\input{figs/experiments/fig_illustrative-example}

\looseness-1\textbf{Without fine-tuning: GPT-4-SS outperforms ~\neurss\,.}
Among our methods that use a base LLM model without fine-tuning, GPT-4-SS achieved the highest scores in all quality attributes across both reference tasks, followed by Llama2-70B-SS (see Figure \ref{fig:result-details}). Additionally, GPT-4-SS performs significantly better than the \textsc{NeurSS} baseline w.r.t. \textsc{Q-overall} in both reference tasks ($p \le 0.05$), based on the $\chi^2$ test \cite{f3e1a971-f77a-3568-83d6-8c0b7f53e9e5}.\footnote{$\chi^2$ tests reported here are computed on aggregated data across both the reference tasks.} \textsc{Q-overall} scores of GPT-3.5-SS and Llama2-7B-SS are lower than that of the baseline \textsc{NeurSS} (which also motivates why we need to do fine-tuning discussed in Section~\ref{sec:fine-tune-llm}).

\textbf{Fine-tuning shows significant improvements.} Our methods using fine-tuning, namely GPT-3.5ft-SS, Llama2-7Bft-SS, and Llama2-70Bft-SS, demonstrate significant improvements compared to using their base versions without fine-tuning ($p \le 0.05$), as shown in Figure \ref{fig:both_stu_and_task}. Remarkably, for HoCMaze-18, there is no significant difference between the performances of GPT-3.5ft-SS and human tutors in \textsc{TutorSS} ($p > 0.05$). We observe that fine-tuning enhances a base LLM's ability to capture the target task's structure (\textsc{Q-task}), as shown in Figure \ref{fig:result-details} -- this improvement is expected given they are fine-tuned to generate solutions for tasks. More importantly, their ability to capture the student's behavior (\textsc{Q-\student}) also increases across all reference tasks and fine-tuned models. Figure \ref{fig:loss} provides insights into the fine-tuning process.

\looseness-1\textbf{Example of synthesized student's attempt.} In Figure~\ref{fig:qualitative-results}, we investigate the scenario for HoCMaze-18 from Figure~\ref{fig:problem-setup}. In this scenario, the student \student's misconception is ignoring conditionals when attempting to solve the given task. Figure~\ref{fig:qualitative-results}(a) shows student code $C^{\student}_{T^{\target}}$ for the target task provided in the benchmark. Figures~\ref{fig:qualitative-results}(b-e) show student codes $\widehat{C}^{\student}_{T^{\target}}$ synthesized by different methods.
Student code synthesized by GPT-3.5ft-SS has the same misconception observed in $C^{\student}_{T^{\reference}}$, while adapting to $T^{\target}$ (\textsc{Q-overall }$\!=\!1$). Notably, it is very close to the code written by human tutors in \textsc{TutorSS} (\textsc{Q-overall }$\!=\!1$). Llama2-70Bft-SS synthesized a code that captures the student's misconception, but only partially reflects the target task's characteristics (\textsc{Q-overall }$\!=\!0.5$). In contrast, the \textsc{NeurSS} baseline synthesized a code that overfits $C^{\student}_{T^{\reference}}$, failing to reflect the layout of $T^{\target}$ as it uses \textcode{turnLeft} blocks instead of \textcode{turnRight} blocks (\textsc{Q-overall }$\!=\!0$).

%% file: figs/experiments/fig_results.tex
\setlength{\tabcolsep}{2pt}
\begin{figure}
    \begin{subfigure}{1\textwidth}
        \centering
        \input{figs/experiments/table_results}
        \subcaption{\looseness-01 Detailed results w.r.t. to each attribute in the quality rubric: \textsc{Q-overall}, \textsc{Q-\student}, and \textsc{Q-task}. Fine-tuned models are highlighted in green. Fine-tuning improves LLMs' capabilities of capturing both student's behavior and target task's characteristics.}
        \label{fig:result-details}
    \end{subfigure}
    \vfill
    \begin{subfigure}{1\textwidth}
        \centering
        {\includegraphics[trim=5em 5em 4em 4em, clip, width=0.5\textwidth]{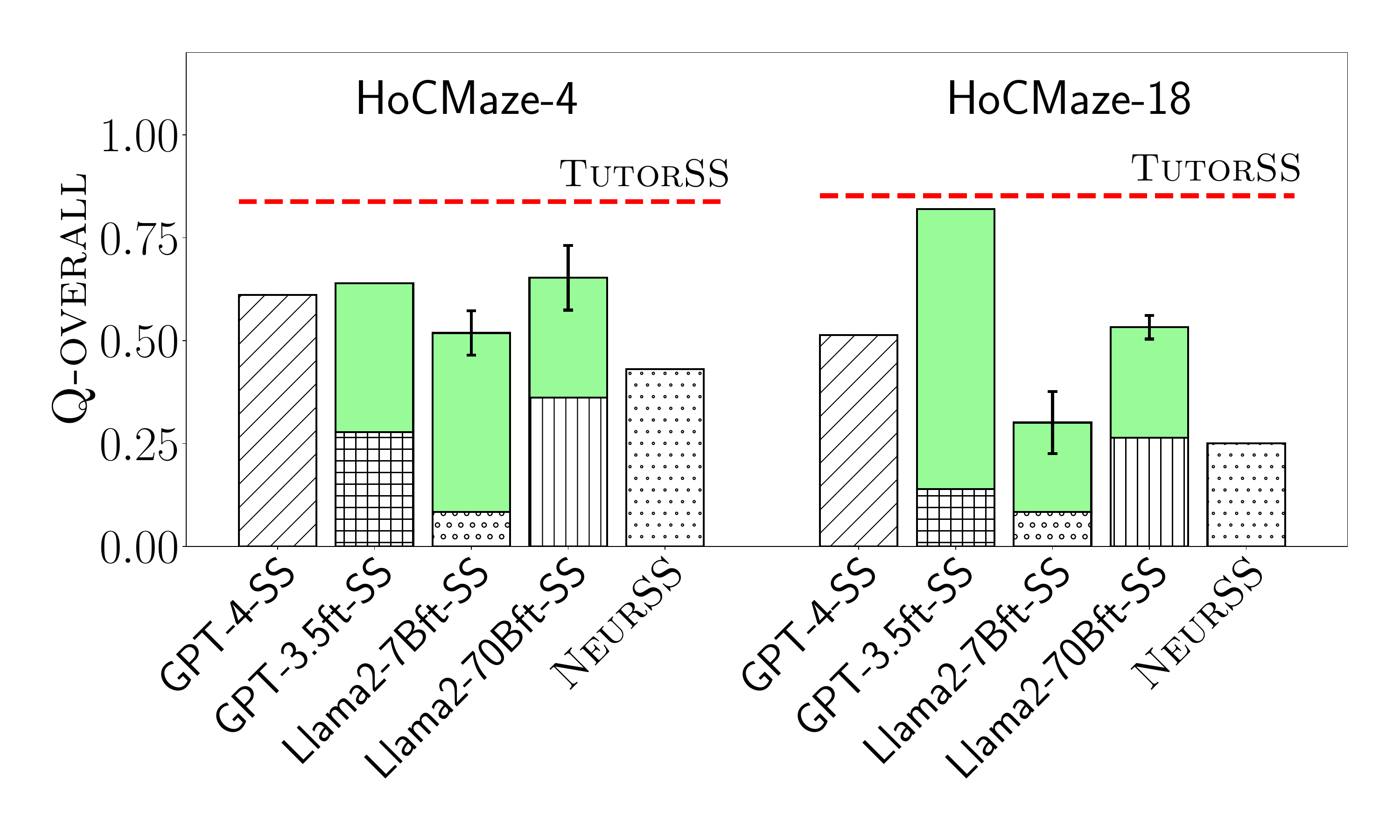}}
        \subcaption{Overall performance (\textsc{Q-overall}). Green areas correspond to fine-tuning improvements. \textsc{TutorSS} (red lines) serves as a performance upper bound.}
        \label{fig:both_stu_and_task}
    \end{subfigure}
     \caption{(a) shows performances of methods w.r.t. individual attributes in our quality rubric. (b) shows overall performance of capturing both student's behavior and target task's characteristics. Human tutors (\textsc{TutorSS}) serve as an oracle. For methods using a fine-tuned LLM, we report numbers averaged over three fine-tuning runs with standard errors (except GPT-3.5ft-SS with only one run, due to the high costs of using fine-tuning APIs from OpenAI).}
     \vspace{-6mm}
    \label{fig:results}
\end{figure}

%% file: figs/experiments/table_results.tex
\scalebox{0.7}{
    \renewcommand{\arraystretch}{1.04}
    \setlength{\tabcolsep}{4pt}
    \begin{tabular}{l||lll|lll}
            \toprule
            & \multicolumn{3}{c|}{\textbf{HoCMaze-4}} & \multicolumn{3}{c}{\textbf{HoCMaze-18}} \\
                & \textsc{Q-overall} & \textsc{Q-\student} & \textsc{Q-task} & \textsc{Q-overall} & \textsc{Q-\student} & \textsc{Q-task} \\ 
            \midrule
            GPT-3.5-SS & $0.28$ & $0.56$ & $0.50$ & $0.14$ & $0.61$ & $0.25$ \\
            GPT-4-SS   & $0.61$ & $0.86$ & $0.72$ & $0.51$ & $0.81$ & $0.58$ \\
            \rowcolor{\oursrowcolor}
            GPT-3.5ft-SS & $0.64$ & $0.69$ & $0.75$ & $0.82$ & $0.92$ & $0.86$ \\ 
            \hline
            Llama2-7B-SS & $0.08$ & $0.14$ & $0.44$ & $0.08$ & $0.25$ & $0.39$ \\
            Llama2-70B-SS & $0.36$ & $0.58$ & $0.50$ & $0.26$ & $0.56$ & $0.50$ \\
            \rowcolor{\oursrowcolor}
            Llama2-7Bft-SS  & $0.52\ (0.05)$      & $0.55\ (0.07)$      & $0.90\ (0.05)$ & $0.30\ (0.08)$ & $0.66\ (0.11)$ & $0.39\ (0.09)$  \\
            \rowcolor{\oursrowcolor}
            Llama2-70Bft-SS & $0.65\ (0.08)$ & $0.87\ (0.05)$ & $0.73\ (0.05)$     & $0.53\ (0.03)$ & $0.83\ (0.02)$ & $0.63\ (0.03)$  \\ 
            \hline
            \textsc{NeurSS} & $0.43$ & $0.56$ & $0.67$ & $0.25$ & $0.78$ & $0.36$ \\
            \textsc{TutorSS} & $0.84$ & $0.92$ & $0.89$ & $0.85$ & $0.89$ & $0.95$ \\
            \bottomrule
    \end{tabular}
}

%% file: figs/experiments/fig_loss.tex
\begin{figure*}[t!]
    \centering
    \includegraphics[trim=3em 3em 3em 18em, clip, width=0.9\textwidth]{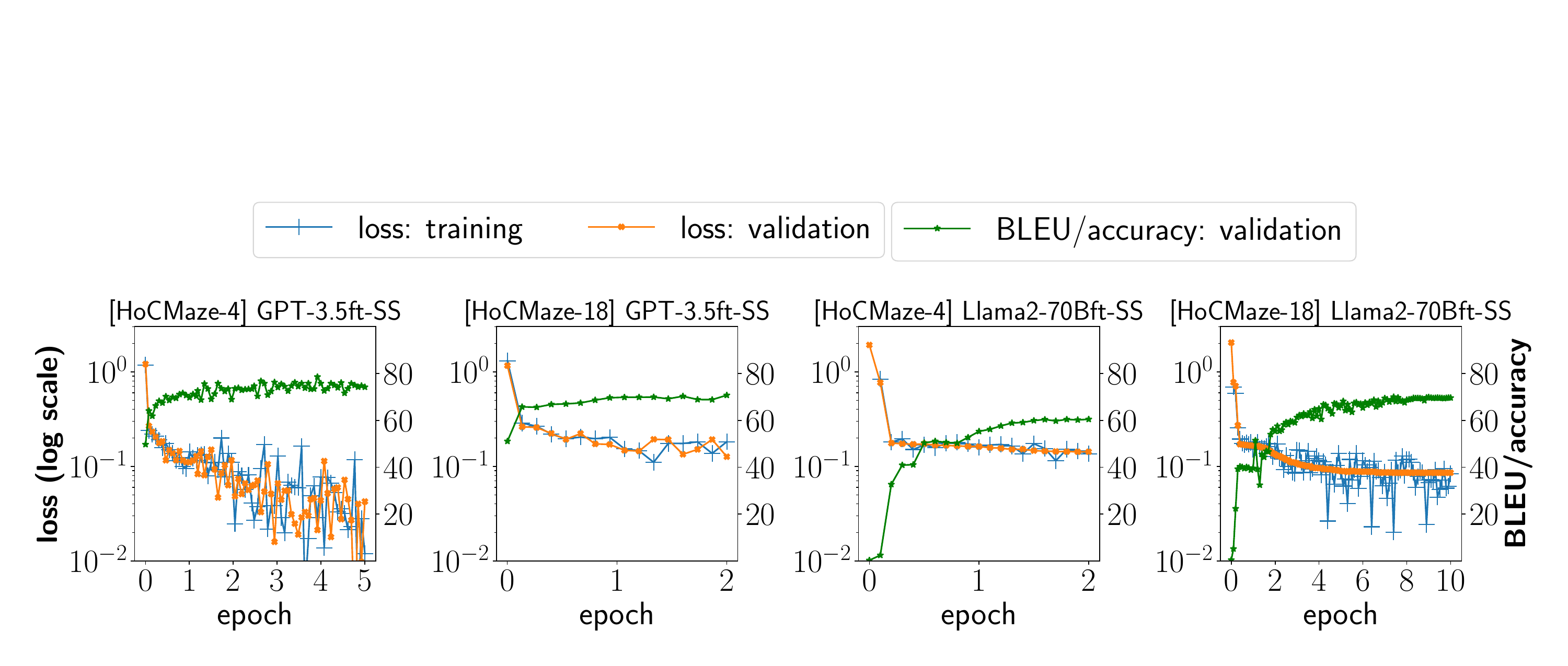}
    \caption{\looseness-01 Losses and evaluations during fine-tuning our two best-performing methods GPT-3.5ft-SS and Llama2-70Bft-SS. We plot data per 0.1 epoch. Losses are plotted on log-scale for better visibility of dynamics. Validation BLEU/accuracy metrics are decided by the fine-tuning library/platform and shown as a sanity check, and are not used for optimization. For GPT-3.5ft-SS, the number of epochs depends on budget spent for OpenAI APIs; we spent roughly half of the total budget for each task. For Llama2-70Bft-SS, the number of epochs are determined by generative performance on a small validation set of examples.}
    \label{fig:loss}

\end{figure*}

%% file: figs/experiments/fig_illustrative-example.tex
\begin{figure*}
    \centering
    \includegraphics[trim=1em 0em 0.1em 2.5em, clip, width=0.98\textwidth]{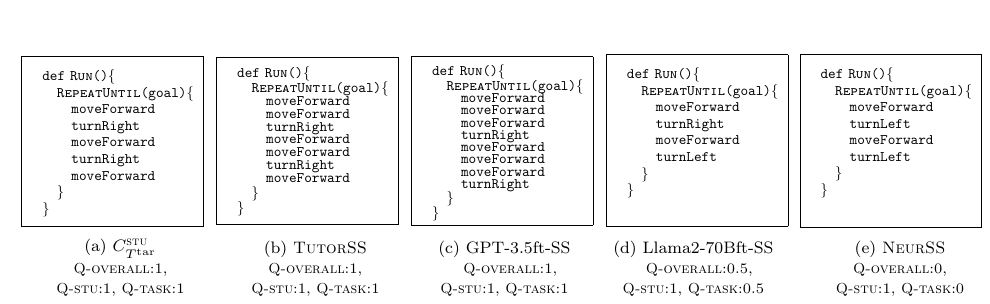}
    \caption{Student \student's attempts for the scenario shown in Figure \ref{fig:problem-setup}. (a) shows ground-truth student \student's attempt $C^{\student}_{T^{\target}}$ provided in the \textsc{StudentSyn} benchmark. (b-e) show synthesized student \student's attempts $\widehat{C}^{\student}_{T^{\target}}$  provided by different methods.}
    \label{fig:qualitative-results}
\end{figure*}

%% file: 6_conclusions.tex

\section{Concluding Discussions}
We proposed a novel LLM-based framework, \textsc{LLM-SS}, for in-context student modeling in open-ended learning domains. The results showcase that methods instantiated from \textsc{LLM-SS} are capable of modeling a student's observed behavior and synthesizing the student's attempt on a target task. We also highlight that fine-tuning a base LLM using expert knowledge in a given open-ended learning domain significantly improves its effectiveness in student modeling. More importantly, our framework does not require building a complex training pipeline as existing works, making it broadly applicable to new domains. In summary, our work demonstrates the potential of using LLMs for in-context student modeling, especially in challenging open-ended learning domains.

Next, we discuss some limitations of our current work and ideas to tackle them in the future. First, our framework was evaluated on one visual programming domain, and the scenarios we considered do not fully capture the wide spectrum of open-ended learning domains; it would be interesting to evaluate our framework in other open-ended learning domains (e.g., algebra or text-based programming). Moreover, it would also be useful to do a more systematic analysis to see which misconceptions or students' behaviors are not well captured by our framework. Second, we provided a student's context through only one example of a problem-solving attempt; it would be interesting to evaluate the effectiveness of our framework when the student's context contains richer information, including the student's background and attempts on different tasks. Third, we evaluated our framework on student modeling metrics but have not evaluated how this modeling helps improve the performance of downstream applications; as future work, it would also be important to investigate the usefulness of our modeling framework directly in downstream applications, such as performance prediction, task recommendation, or synthetic behavioral dataset generation for training data-intensive models. In particular, since our framework allows fine-grained synthesis of a student's attempts beyond binary performance prediction, it would be interesting to see how our framework can potentially be applied for providing finer-grained feedback to the student about possible misconceptions. 

Finally, we note that there are several ethical implications regarding the use of LLMs for student modeling. For instance, the attempts synthesized by LLMs may not accurately reflect a student's understanding or ability. Moreover, LLMs are prone to hallucination and might generate inaccurate information. Therefore, it is crucial to implement appropriate validation mechanisms and safeguards when deploying LLM-based student modeling techniques in classrooms.

%% file: 7_acknowledgements.tex
\begin{ack}
Funded/Co-funded by the European Union (ERC, TOPS, 101039090). Views and opinions expressed are however those of the author(s) only and do not necessarily reflect those of the European Union or the European Research Council. Neither the European Union nor the granting authority can be held responsible for them.
\end{ack}